\documentclass[letterpaper, conference]{ieeeconf}

\IEEEoverridecommandlockouts  

\usepackage{graphicx}
\usepackage{subcaption}
\usepackage{amsmath}
\usepackage{amsmath}
\usepackage{amssymb}
\usepackage{cite}
\usepackage{xcolor}
\usepackage{booktabs}
\usepackage{multirow}
\usepackage{tabularx}

\title{\LARGE \bf
Learning to Walk in Costume: Adversarial Motion Priors for Aesthetically Constrained Humanoids}

\author{
    Arturo Flores Alvarez${^1}$,
    Fatemeh Zargarbashi${^2}$,
    Havel Liu${^1}$,
    Shiqi Wang${^1}$,\\
    Liam Edwards${^1}$,
    Jessica Anz${^1}$,
    Alex Xu${^1}$,
    Fan Shi${^3}$,
    Stelian Coros${^2}$,
    Dennis W.~Hong${^1}$\\
    \thanks{${^1}$ Robotics and Mechanisms Laboratory (RoMeLa), Dept. of Mechanical and Aerospace Engineering, UCLA, CA, USA.}
    \thanks{${^2}$ Computational Robotics Lab (CRL), Dept. of Computer Science, ETH Zürich, Switzerland.}
    \thanks{${^3}$ Human‑Centered Robotics Lab, Dept. of Electrical and Computer Engineering, National University of Singapore, Singapore.}
    \thanks{
    The authors thank Min Sung Ahn for his expertise in sim-to-real transfer and best practices on hardware, Norman Zhu for sharing model-based walking experimental data on Cosmo, Conrad Ku and Premkumar Sivakumar for their support during testing phases, and Yusuke Tanaka and Alvin Zhu for their insights on improved training methodologies.
    }
}

\begin{document}

\maketitle

\begin{abstract}
We present a Reinforcement Learning (RL)-based locomotion system for Cosmo, a custom-built humanoid robot designed for entertainment applications. Unlike traditional humanoids, entertainment robots present unique challenges due to aesthetic-driven design choices. Cosmo embodies these with a disproportionately large head (16\% of total mass), limited sensing, and protective shells that considerably restrict movement. To address these challenges, we apply Adversarial Motion Priors (AMP) to enable the robot to learn natural-looking movements while maintaining physical stability. We develop tailored domain randomization techniques and specialized reward structures to ensure safe sim-to-real, protecting valuable hardware components during deployment. Our experiments demonstrate that AMP generates stable standing and walking behaviors despite Cosmo's extreme mass distribution and movement constraints. These results establish a promising direction for robots that balance aesthetic appeal with functional performance, suggesting that learning-based methods can effectively adapt to aesthetic-driven design constraints. %that would typically challenge traditional control approaches.
\end{abstract}

\section{Introduction}
    
Humanoid robots play an increasingly vital role in human environments thanks to their intuitive interaction capabilities and adaptability.
Their anthropomorphic design equips them to navigate and perform tasks within spaces designed for humans, ranging from simple locomotion to intricate manipulation tasks~\cite{humanoidsurvey2025}.
Particularly, in the entertainment sector, humanoid robots enrich storytelling and audience engagement through lifelike movements, significantly enhancing performances across films, theme parks, and live shows. \cite{toothless, grandia2025design}.
    
Entertainment humanoids face unique challenges: disproportionate body parts shifting center of mass, restricted sensing from aesthetic shells, and limited joint mobility from protective coverings—all prioritizing visual appeal over stability.

Our robot, Cosmo, encompasses these challenges: it lacks an onboard vision system with an elevated center of mass, making it an ideal case study for testing the limits of conventional locomotion methods and exploring the capabilities of learning-based solutions. 

\begin{figure}[t]
    \centering
    \includegraphics[width=.99\linewidth]{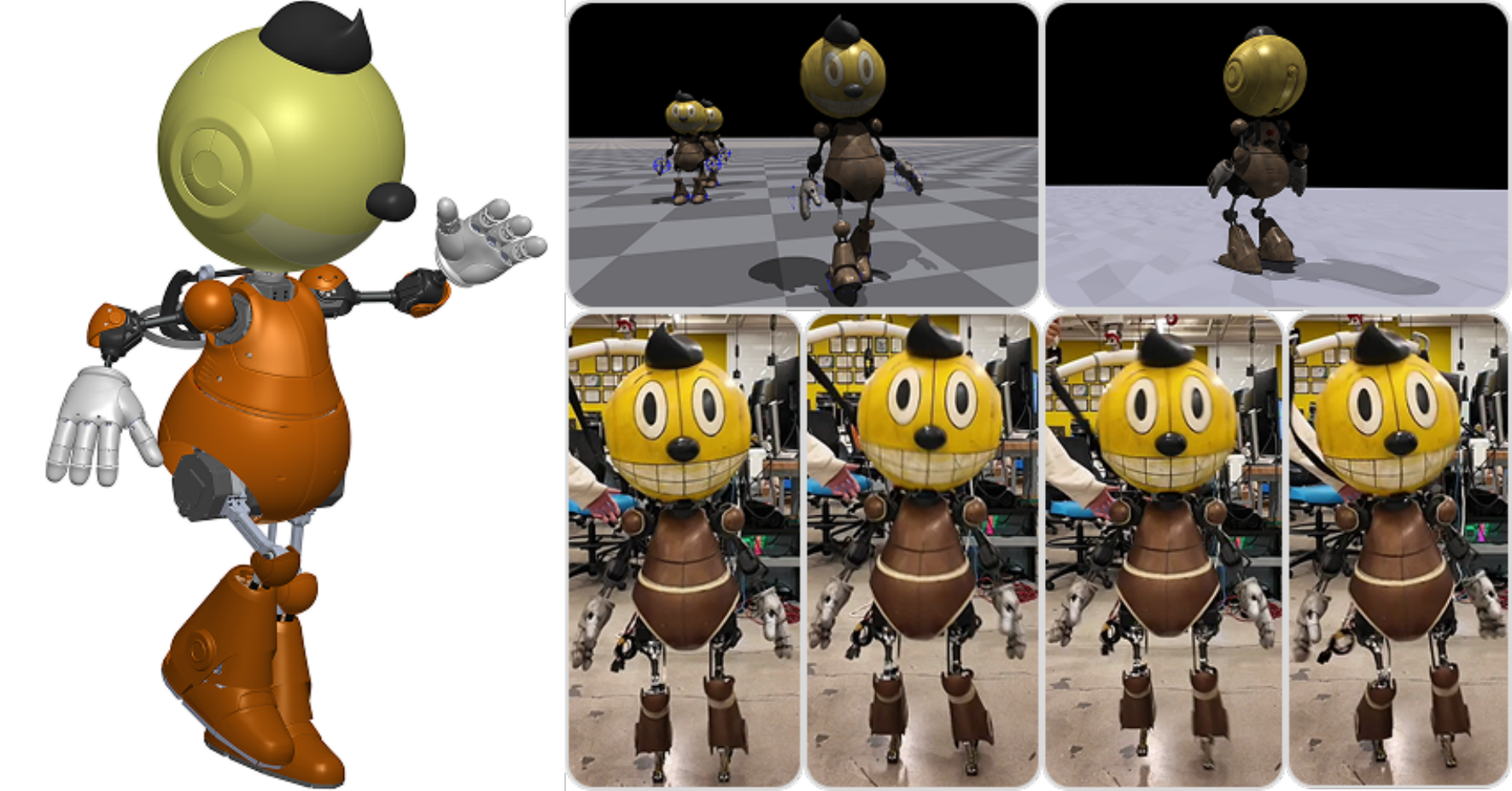}
    \caption{Cosmo: an entertainment humanoid robot with covers designed for a blockbuster movie. (\textbf{Left}): CAD Design. (\textbf{Top}): Using Isaac Gym's massively parallelized environments to train with different styles and terrain. (\textbf{Bottom}): Sim-to-Real demonstration of natural walking (see supplementary video).}
    \label{fig:brek1}
\end{figure}

In this work, we demonstrate how modern learning-based methods, particularly Adversarial Motion Priors (AMP) \cite{peng2021amp}, can overcome these difficulties.
AMP is a method that blends the realism of imitation learning with the flexibility of reinforcement learning, enabling robots to develop physically plausible movements while retaining the signature style of human movements.

We begin by drawing on human motion capture data from established datasets, then retargeting it onto our custom robot character.
This step ensures the resulting movements preserve the essence of human behavior while adapting to the unique physical constraints of our platform. 

By combining AMP with comprehensive sim-to-real strategies—including domain randomization and meticulous hardware parameter tuning—we achieve real-world locomotion on Cosmo, addressing challengs in balancing, walking, and safe deployment. Our key contributions are:

% --- 5. Contribution:
\begin{enumerate}
    \item A learning-based locomotion control system for a humanoid with an extreme mass distribution and shifted center of mass—a configuration rarely addressed in humanoid control literature.
    \item A sim-to-real transfer pipeline tailored for robots with aesthetic shell constraints, enabling safe transfer from simulation to physical hardware while preserving both performance and component safety.
    \item A demonstration that AMP-guided reinforcement learning can generate natural, stable walking behaviors even on platforms with significant mechanical limitations and entertainment-focused design constraints.
\end{enumerate}

This work addresses an emerging challenge in entertainment robotics: achieving stable and expressive locomotion in designs where aesthetics dominate over functionality.
By showcasing how learning-based methods can overcome these unique constraints, we highlight a promising path forward for robots that must balance visual appeal with real-world performance. Our results suggest that learning-based methods, especially when guided by motion priors, offer a powerful tool for enabling dynamic motion in robots destined for rich, interactive entertainment environments.

\section{Related Work}
Model-based approaches have long been the foundation of humanoid locomotion \cite{yamamoto2020survey,kuindersma2016optimization}.
These controllers typically require accurate models and extensive manual tuning, and they are often brittle when conditions deviate from those assumed in the design.
In contrast, reinforcement learning (RL) has emerged as a powerful alternative for humanoid control~\cite{radosavovic2024humanoidRL,Li2024RLbipedal, cheng2024expressive}.
Pure RL approaches, such as~\cite{li2021cassie, zhuang2024parkour, zhang2024wococo}, have demonstrated impressive dynamic capabilities, enabling agile behaviors like running and parkour.
These methods often assume well-structured morphologies and full-state sensing. 
Despite their success, purely RL-based methods typically require extensive reward shaping and often produce unnatural, non-human-like movements due to the absence of imitation signals.
Imitation learning offers a way to imbue humanoid controllers with human-like movement qualities by learning from demonstration data (such as motion capture of human locomotion). A significant practical advantage of this approach is that it establishes clear visual benchmarks for expected behavior—when the robot deviates from reference motions, these discrepancies can be quickly identified and debugged through visual inspection.
In particular, Peng et al.~\cite{peng2021amp} introduced Adversarial Motion Priors (AMP) to incorporate motion data into RL training loop via adversarial training, producing natural walking and running motions. ~\cite{escontrela2022adversarial}  further demonstrated that AMP can replace complex reward engineering by learning natural motion behaviors through  adversarial training alone.

Our work builds upon these methods but differs in two key ways: (1) our hardware is significantly more difficult to control due to its top-heavy design, and (2) our system lacks visual input, forcing the policy to rely entirely on proprioception. While imitation learning and domain randomization are well-studied in sim-to-real transfer, our results demonstrate their efficacy under harder conditions.

\begin{figure}[t]
    \centering
    \includegraphics[width=1.0\linewidth]{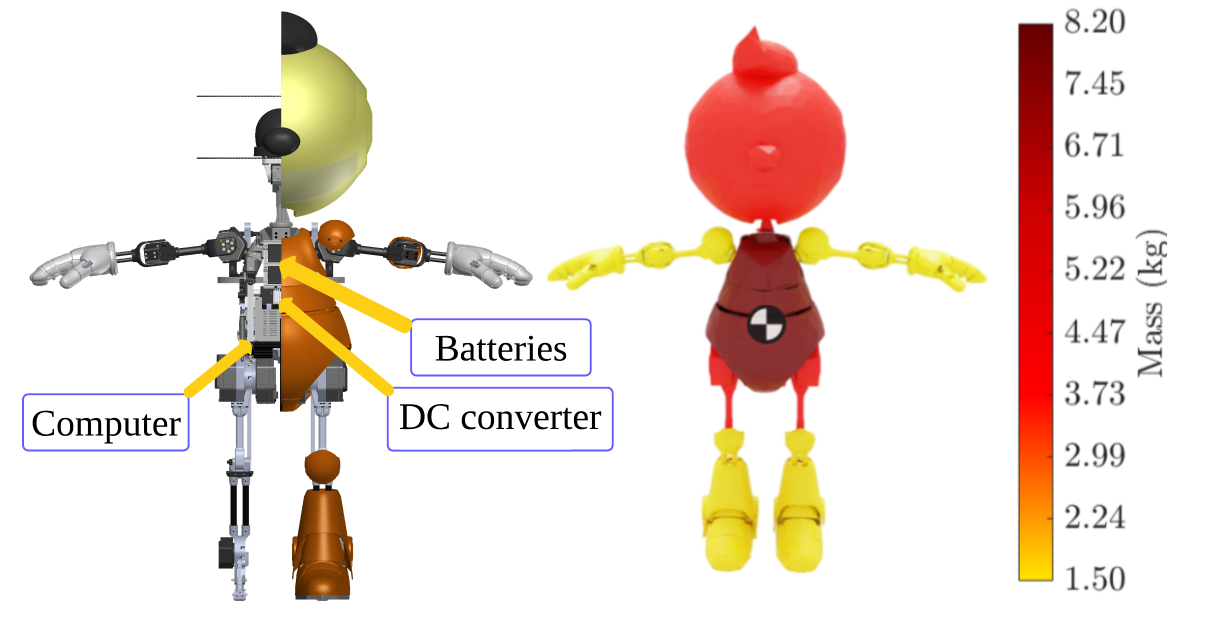}
    \caption{Cosmo visualization: (left) arm range of motion and internal vs. exterior housing comparison; (right) mass distribution analysis highlighting the disproportionate head mass.}
    \label{fig:cosmo_vitruvian_mass}
\end{figure}

\section{Methods}
\label{sec:methods}

\begin{figure*}[t]
    \centering
    \includegraphics[width=0.85\linewidth]{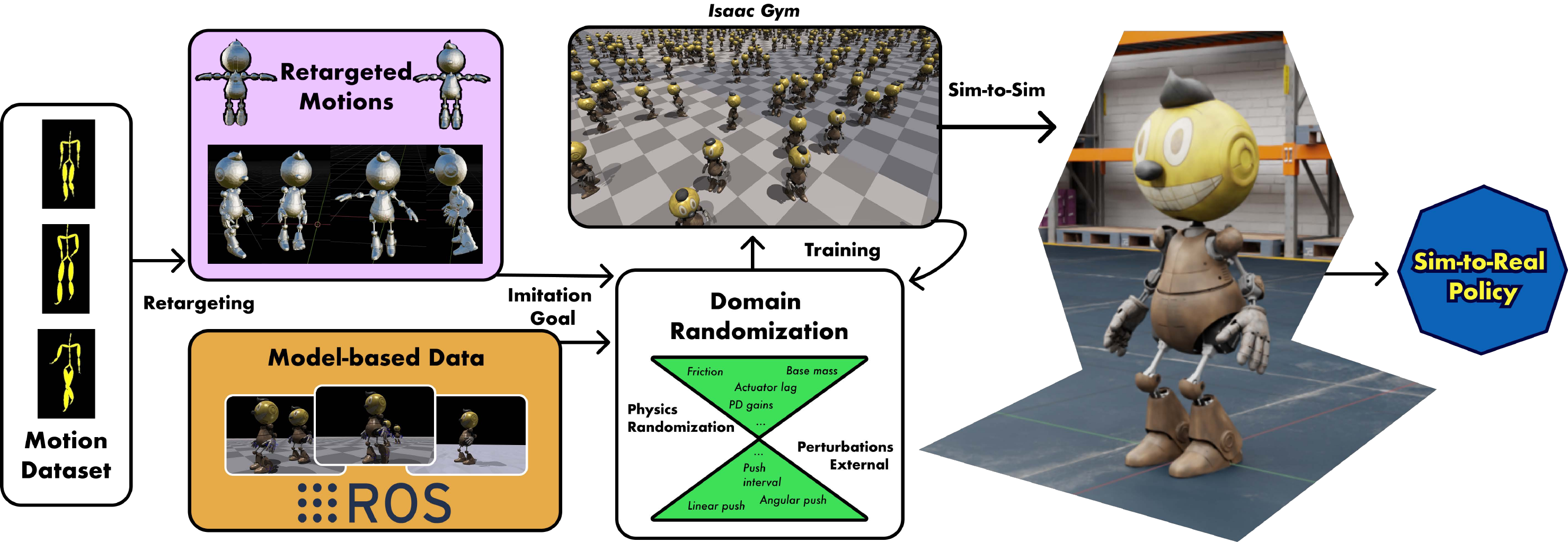}
    \caption{Sim-to-Real pipeline: (a) Retargeting from diverse data sources, (b) Training, (c) Validation, (d) Deployment}
    \label{fig:pipeline}
\end{figure*}

\subsection{Motion Retargeting}
The increasing availability of human motion capture data has become a natural and effective choice for generating reference motions for lifelike robots. However, a significant challenge arises due to differences in morphology. For example, Cosmo's legs rotate at two points in the hip, whereas a human just has a ball joint at the hip. Therefore, an intermediate process is necessary to adapt human motion data to the kinematic constraints of a specific robotic platform. This process is commonly known as \emph{retargeting}.

We utilize the Rokoko plugin \cite{rokoko_blender_plugin} in Blender \cite{blender} for retargeting CMU Mocap Dataset \cite{cmumocap,fbx_mocap}. A custom animation rig matching the Cosmo robot’s proportions is built in Blender. Human motion data is approximated by averaging limb movements to fit this rig. Additionally, the retargeting does not consider the wide shells on Cosmo. This is especially prevalent in the foot shells and results in the meshes clipping into each other.

\subsection{Imitation Learning with AMP}
We frame the locomotion problem as a Partially Observable Markov Decision Process (POMDP) where the agent must learn a policy $\pi(a|s)$ that maps observations $s$ to actions $a$. The training objective maximizes the expected discounted return:
\begin{equation}
J(\pi) = \mathbb{E}_{\tau \sim \pi} \left[ \sum_{t=0}^{T} \gamma^t r(s_t, a_t) \right]    
\end{equation}

\begin{figure}[t]
    \centering
    \includegraphics[width=1.0\linewidth]{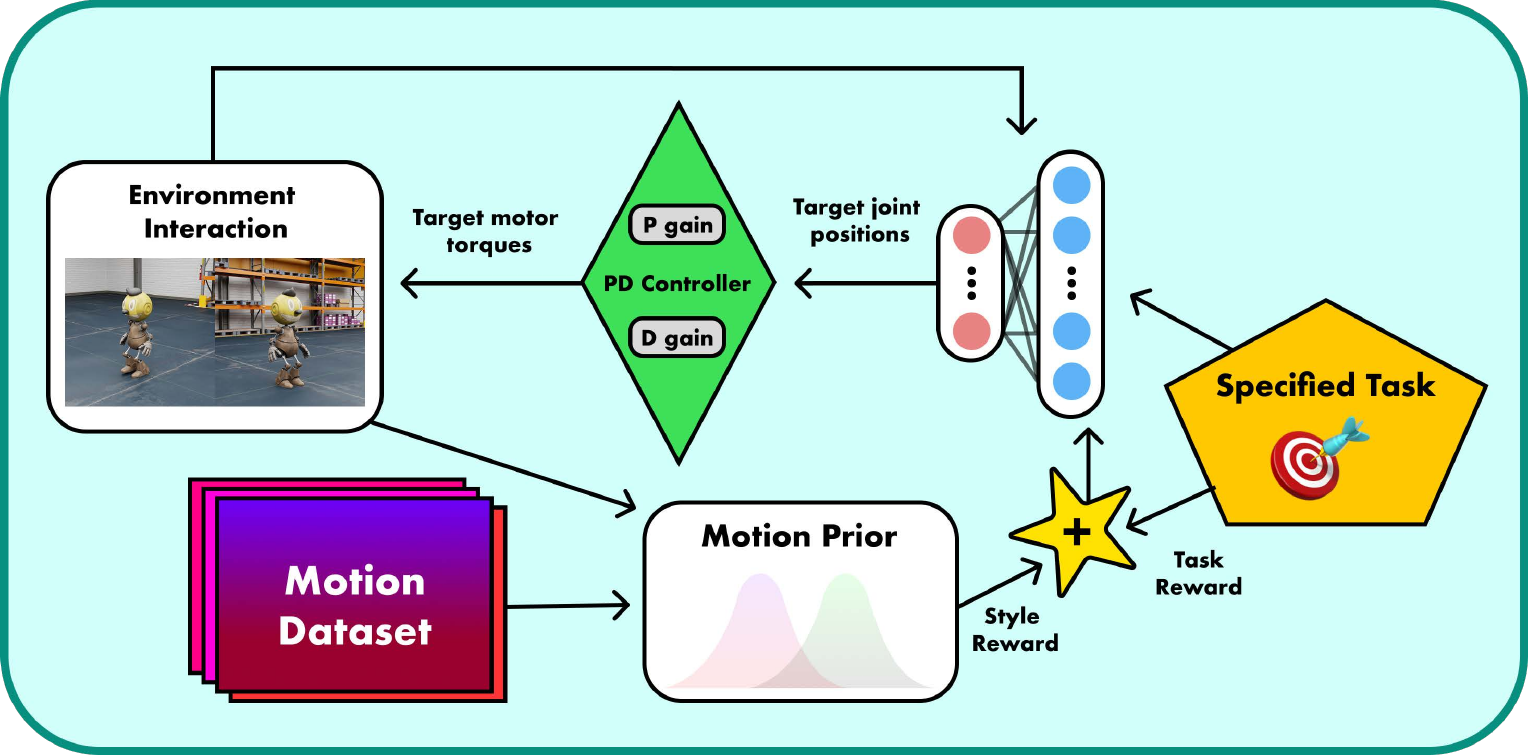}
    \caption{Schematic overview of the training framework.}
    \label{fig:trainingframework1}
\end{figure}

Our approach utilizes Adversarial Motion Priors (AMP) \cite{peng2021amp} to learn locomotion from human motion capture clips. Figure \ref{fig:trainingframework1} provides an overview of our complete training framework.
AMP incorporates a discriminator network $D_\phi(s)$ that distinguishes motions from the reference dataset $\mathcal{M}$ from states generated by the policy $\pi_\theta$, providing a learned reward signal that encourages lifelike motion. 
% This methodology effectively blends imitation learning with reinforcement learning, encouraging physically stable yet naturally expressive behaviors.

% The core of the AMP approach relies on an adversarial discriminator network $D_\phi(s)$ that distinguishes between the policy-generated states and reference motion data. This discriminator provides a learned reward signal that guides the policy toward generating human-like motion:
The loss function for the discriminator is
\begin{align}
    \mathcal{L}_D(\phi) = &-\mathbb{E}_{s_t \sim \pi_\theta}[\log(1-D_\phi(s_t))] \nonumber \\
    &- \mathbb{E}_{s_t^{ref} \sim \mathcal{M}}[\log(D_\phi(s_t^{ref}))].
\end{align}

The policy then incorporates the discriminator's output as a reward term $r_{AMP}(s_t) = \log D_\phi(s_t)$, encouraging the agent to generate motions that appear natural while simultaneously satisfying the task objectives and physical constraints. The discriminator is trained alongside the policy in an adversarial manner, continuously adapting to distinguish increasingly realistic policy behaviors. The policy network uses 3 layers (512, 256, 128 units), critic and discriminator use 2 layers (256, 128 units), all with ELU activations.

\textbf{Observation space}
Our observation is a state vector $s \in \mathbb{R}^{d_s}$ comprising proprioceptive information acquired from the motor chain and onboard  state estimator:
% This state representation is mathematically formulated as:
\[
s = [v_{\text{base}}; \, \omega_{\text{base}}; \, q - q_{\text{default}}; \, \dot{q}; \, g_{\text{proj}}; \, a_{\text{prev}}; \, h_{\text{base}}; \, c_{\text{cmd}}].
\]
This formulation captures the robot's complete dynamic state through base linear velocity ($v_{\text{base}} \in \mathbb{R}^3$), angular velocity ($\omega_{\text{base}} \in \mathbb{R}^3$), normalized joint positions ($q - q_{\text{default}} \in \mathbb{R}^{n_j}$), joint velocities ($\dot{q} \in \mathbb{R}^{n_j}$), projected gravity orientation ($g_{\text{proj}} \in \mathbb{R}^3$), previous actions ($a_{\text{prev}} \in \mathbb{R}^{n_j}$), base height ($h_{\text{base}} \in \mathbb{R}$), and command signals ($c_{\text{cmd}} \in \mathbb{R}^3$). Base kinematics are estimated via an invariant extended Kalman filter using proprioceptive data.

The command ranges in Table \ref{tab:command_ranges} reflect the typical operating range of Cosmo, with forward velocity having an asymmetric range to accommodate the natural bias toward forward locomotion. 

\begin{table}[h!]
\centering
\caption{Commanded velocity components in $c_{\text{cmd}}$}
\begin{tabular}{lcc}
\toprule
 \textbf{Description} & \textbf{Command Range} & \textbf{Units}  \\
\midrule
Desired forward velocity & $[-0.3, 0.9]$ & m/s \\
Desired lateral  velocity & $[-0.3, 0.3]$ & m/s \\
Desired yaw rate & $[-0.3, 0.3]$ & rad/s \\
\bottomrule
\end{tabular}
\label{tab:command_ranges}
\end{table}

\textbf{Action space}
The action space consists of target joint positions for all actuated degrees of freedom. These target positions are converted to torques through a PD controller running at a higher frequency than the policy. 

\textbf{Reward Scheme}
Our reward function combines task-oriented terms with the style reward from AMP. The task rewards are organized into distinct groups, as shown in Table \ref{tab:reward_designs}, with adjustable coefficients to enable curriculum learning.

\begin{table}[h]
\centering
\caption{Reward Components for AMP Control}
\label{tab:reward_designs}
\begin{tabular}{cc}
\hline
\textbf{Components} & \textbf{Formula} \\
\hline
 \multicolumn{2}{c}{\textbf{Imitation}}  \\
\hline
AMP Reward & $-\log D_\phi(s)$  \\
\hline
 \multicolumn{2}{c}{\textbf{Motion Quality}} \\
\hline
Joint rate & $\exp(- (\dot{q} - \dot{q}_{\text{target}})^2 \sigma^2)$ \\
\hline
 \multicolumn{2}{c}{\textbf{Safety}} \\
\hline
Foot stumble & $\exp(- (F_{\text{max}} - F)^2 / \sigma^2)$ \\
Foot orientation & $r = \exp(- \|\vec{n}_{\text{feet}} - \vec{n}_{\text{ref}}\|^2 / \sigma^2)$  \\
Foot height & $r = \exp(- (h_{\text{feet}} - h_{\text{ref}})^2 / \sigma^2)$ \\
\hline
 \multicolumn{2}{c}{\textbf{Task}}   \\
\hline
Linear velocity & $\exp(- (v_{\text{cmd}} - v_{\text{base}})^2 / \sigma^2)$   \\
Angular velocity & $\exp(- (\omega_{\text{cmd}} - \omega_{\text{base}})^2 / \sigma^2)$   \\
\hline
\end{tabular}
\end{table}

\begin{figure}[ht!]
    \begin{subfigure}[b]{0.195 \textwidth}
        
        \includegraphics[width=\textwidth]{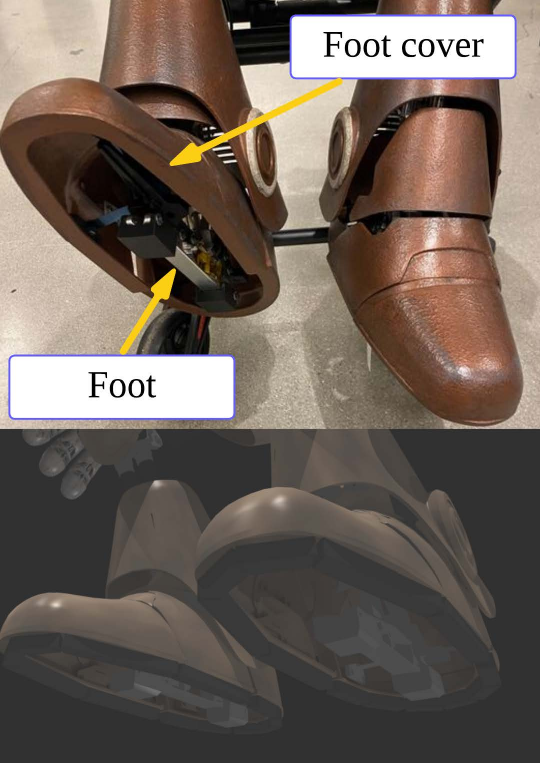}
        \caption{}
        \label{fig:cosmo_feet_covers}
    \end{subfigure}
    \hfill
    \begin{subfigure}[b]{0.28\textwidth}
        
        \includegraphics[width=\textwidth]{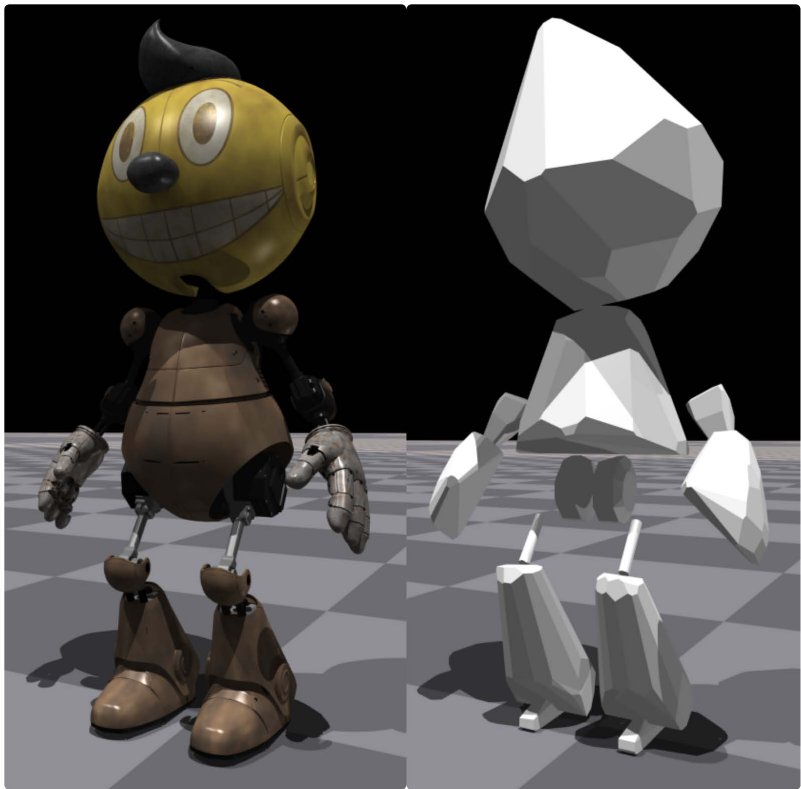}
        \caption{}
        \label{fig:cosmo_no_mb}
    \end{subfigure}
    \caption{(a) Real vs. simulated feet showing protective aesthetic covers. (b) Cosmo's collision meshes and foot model.}
    \label{fig:cosmo_collisions}
\end{figure}

% Given that COSMO represents both a high-value hardware platform and features expensive custom shells, 
We carefully designed reward groups to ensure motion quality and safety while accomplishing the locomotion task. The reward structure comprises three key functional areas and later will be used as metric to compare the performance of different experiments:
\begin{enumerate}
    \item The \textit{Motion Quality} rewards limit joint rate changes to ensure smooth rather than jerky movements. Joint target rates are computed as the L2 norm of consecutive joint target differences, scaled by the control frequency.
    \item The \textit{Safety} rewards protect the robot's aesthetic shells and optimize foot interactions from aggressive impacts. For example, our foot orientation reward specifically keeps feet flat to the plane, preventing the brown protective covers (Figure \ref{fig:cosmo_feet_covers}) from contacting the floor at damaging angles.
    \item The \textit{Task Reward} components directly address the locomotion objectives.
\end{enumerate}

Our collision modeling used simplified convex hulls to represent shells during training while accounting for motion constraints—particularly around the feet where clearance is minimal, and for internal motors that limit yaw rate due to proximity.

Ablation studies in Section \ref{sec:ablation} demonstrate each reward group's critical importance. This balanced reward approach, combined with our simulation-first methodology, minimized physical testing risks and identified potential issues before deployment.

\subsection{Hardware Platform}
Our custom humanoid platform, Cosmo, embodies a fictional character from a blockbuster movie while delivering robust locomotion capabilities. Cosmo's design presents unique control challenges with its disproportionately large head weighing 4 kg—16\% of the 25 kg total mass—creating a top-heavy distribution rarely addressed in humanoid locomotion research (Figure \ref{fig:cosmo_vitruvian_mass}).

% Computation runs on an Intel i7-13700H mini PC, with all current control algorithms executed on CPU, though the architecture supports future GPU parallelization. 
The robot relies exclusively on proprioceptive feedback from its 28 degrees of freedom: 10 in the legs, 8 in the arms, 2 in the head, and 8 in the hands.

For locomotion control, Cosmo employs Westwood Robotics actuators—Panda Bear Plus models for high-torque hip pitch and knee joints, and Koala Bear Muscle Build variants for other primary joints—enabling precise torque control through internal sensing alone.

The design effectively balances aesthetic requirements with functional engineering within severely constrained packaging volumes. The character-defining large head incorporates carbon fiber reinforcement to minimize weight while maintaining visual presence. The feet include spring steel flexures, providing compliance to absorb impact from uneven terrain.

Strategic joint design optimizes mass distribution by positioning actuators close to the torso and minimizing limb inertia. These engineering solutions enable Cosmo to achieve stable locomotion despite its entertainment-first morphology with significantly elevated center of mass.

\subsection{Simulation Suite}
\label{sec:simsuite}
To systematically address the particular instability caused by Cosmo's head, we developed a comprehensive simulation environment in NVIDIA's Isaac Sim \cite{isaacsim} before physical deployment. The physics engine enabled detailed analysis of both environmental and self-collisions, critical for evaluating stability with the disproportionately massive head. For stability assessment, we defined the center of mass, projected gravity vector, and kinematically feasible poses. We constructed a support polygon as the convex hull of foot-ground contact points, with stability determined by whether the gravity vector remained within this polygon during motion transitions. Our methodology involved systematically testing predefined poses by sweeping each joint through its range of motion while monitoring for collisions and stability violations.

We validated these poses on the physical platform to establish sim-to-real correspondence and determine the optimal stance for Cosmo. Comparative analysis against a biomechanically optimized robot ARTEMIS \cite{zhu_design_2023} quantitatively demonstrated Cosmo's inherent instability challenges, revealing center of mass deviations five times greater during equivalent motions.

After stability analysis, we conducted sim-to-sim transfer in Isaac Sim to validate policies safely before hardware deployment. This enabled joint gain tuning without endangering the physical robot and testing of controller robustness against external disturbances and varying surface friction.

For massive parallel training, we employ our pipeline in Isaac Gym \cite{makoviychuk2021isaacgym, massive_rl}, NVIDIA's high-performance physics simulation platform.

\textbf{Push-Recovery/ Balancing Policy}
Based on the pose stability analysis, we implemented an intermediate policy development stage to address two critical concerns: safeguarding the robot's delicate components and acquiring valuable practical insights for more complex motion control. 
The push-recovery policy served as an intuitive debugging case with well-defined expected behaviors.

For implementation, we utilized a concise motion clip featuring the statically stable pose identified in our stability analysis, maintained at a fixed position. 
We trained a push-recovery policy by applying random external forces, allowing the system to learn corrective actions.

This intermediate policy helped identify the most sensitive parameters of our platform—findings that were subsequently validated in our experimental results. 
Critically, this approach also facilitated efficient parameter tuning when transitioning between our two Cosmo platforms, as the push-recovery framework provided a controlled environment to calibrate platform-specific differences while minimizing risk to hardware.

\textbf{Walking Policy}
For our walking policy, we combined diverse motion references to achieve both expressiveness and stability. While we began with motion capture trajectories including "swaggy" walking and standing poses, this data alone proved insufficient for safe deployment—lacking robot-specific dynamics information and offering only fixed velocities typical in mocap datasets. To address these limitations, we supplemented our references with experimental motion data from a hierarchical quadratic programming based inverse dynamics whole-body controller running on Cosmo (Figure \ref{fig:pipeline}), which provided the necessary dynamics information and a continuous spectrum of velocities as commanded. This hybrid approach created a natural curriculum, starting with controlled, lower-speed movements before progressing to more expressive gaits (typically 0.5-0.7 m/s), enabling safe hardware testing while preserving human-like expressiveness.

\subsection{Sim-to-Real Transfer}
Our hardware implementation employs low-level PD control for policy-to-actuation translation:
\[
\tau = K_p (q_{\text{target}} - q) + K_d (\dot{q}_{\text{target}} - \dot{q})
\]
We established a gain-tuning protocol focused on joint tracking, vibration management, and torque constraints, with more aggressive tuning for torso-related joints to manage the robot's disproportionate head mass.

To bridge the reality gap, we implemented domain randomization (Table \ref{tab:domain_rand}) through two key strategies: 1) dynamics randomization of physical parameters (mass, gains, friction, actuator response), and 2) calibrated sensor noise injection to develop policy robustness. We also incorporated periodic external perturbations with tuned magnitudes and intervals.

\begin{table}[!t]
\centering
\caption{Domain Randomization Parameters}
\label{tab:domain_rand}
\begin{tabularx}{\columnwidth}{@{} l X c @{}}
\toprule
\textbf{Parameter} & \textbf{Description} & \textbf{Range} \\
\midrule
\multicolumn{3}{l}{\textit{Physics Randomization}} \\
\midrule
Friction & Contact friction coefficient & [0.2, 1.1] \\
Base mass & Robot base mass perturbation & $\pm$1.5 kg \\
PD gains & Joint PD controller gain multipliers & [0.75, 1.13] \\
Actuator lag & Control signal delay & 4 timesteps \\
\midrule
\multicolumn{3}{l}{\textit{External Perturbations}} \\
\midrule
Push interval & Time between successive external forces & 4.0 s \\
Linear push & Maximum linear push velocity & 0.5 m/s \\
Angular push & Maximum angular push velocity & 0.2 rad/s \\
\bottomrule
\end{tabularx}
\end{table}

\section{Experimental Results}
\label{sec:results}
Despite the robot's challenging morphology and constrained range of motion, our trained policy demonstrates stable standing and walking on flat surfaces. 
In this section, we outline the primary challenges encountered during the training process, simulation, and hardware deployment, along with the strategies employed to address them.
Our approach involved an extensive study of the platform using cutting-edge simulators followed by numerous hardware trials.
A comprehensive justification of our design and methodological choices is provided in the ablation studies section, where we further analyze the specific difficulties encountered and the effectiveness of our solutions.

\begin{figure}[t]
    \centering
    \includegraphics[width=1.0\linewidth]{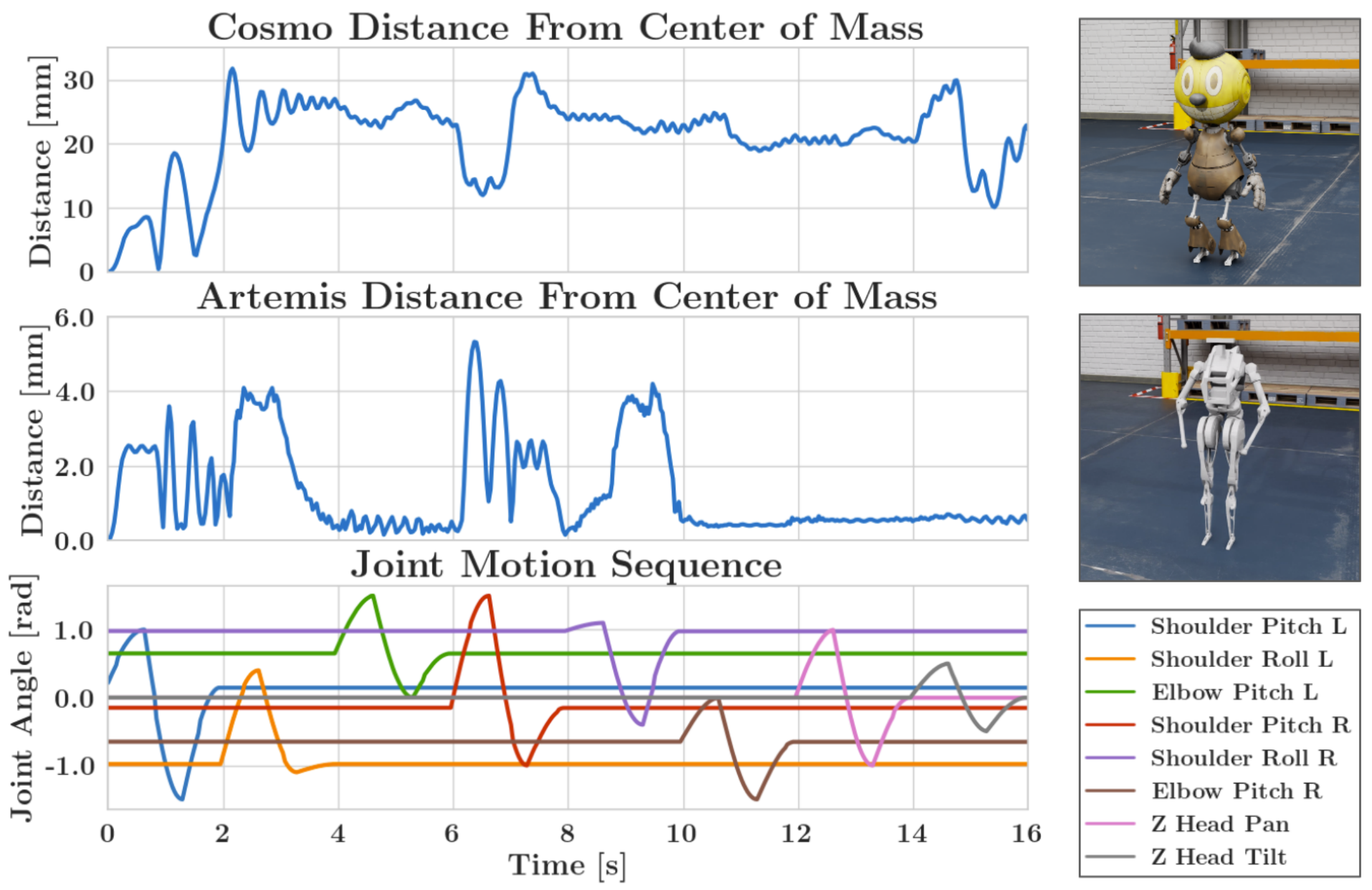}
    \caption{Pose stability testing tracking the center mass disturbance from the initial center of mass position for stable poses. (\textbf{Top}) test results on Cosmo, (\textbf{Middle}) test results on a traditional humanoid ARTEMIS, (\textbf{Bottom}) the position of the all joints during test, causing center of mass disturbances. Both robots share identical arm configurations, so joint correspondences are equivalent.}
    \label{fig:nn1}
\end{figure}

\subsection{Pose Stability Analysis and range of motion test (Isaac Sim)}

\textbf{Static Analysis} To identify stable configurations, we evaluated four poses derived from a standard human standing stance with arms at the sides. Using the projected gravity vector relative to the foot-defined stability region, we assessed pose viability during motion. In simulation, only 2 of 4 poses maintained stability throughout their joint range of motion, while the others resulted in falls when the gravity vector exited the support region. Additional results for tested poses are included in the paper's corresponding video.

Figure \ref{fig:nn1} illustrates this analysis and our selected initial pose. The top plot tracks the Euclidean distance of the center of mass from its initial position during joint motion in a stable pose. For comparison, the middle plot shows identical testing on ARTEMIS. The results are striking: Cosmo's center of mass deviated by 30 mm—five times more than ARTEMIS (6 mm)—quantitatively confirming our entertainment-focused design's inherent instability challenges. The bottom plot displays the joint motion sequence used in testing to cause the devitations. For example, the movement of the right shoulder pitch joint corresponds to a large disturbance in both models, which matches intuition that forward motion of the arm limb would cause disturbances to stability. This analysis directly informed our initial pose selection for control development and training, establishing a practical starting point for managing the robot's challenging mass distribution.f

\subsection{Training results (Isaac Gym)}
We identified optimal training parameters using Random State Initialization across different starting conditions. For performance evaluation, we used two key metrics: reward components (where values closer to 1.0 indicate better performance) and AMP discriminator loss (where values closer to 0 indicate better style matching with reference motions).

\subsubsection{Balancing}
Training proved highly sensitive to the robot's head mass distribution. We systematically evaluated balancing policy performance by varying head mass in the robot's URDF file while maintaining consistent inertial distribution and head randomization parameters, as shown in Table \ref{tab:performance_metrics}.

Our findings reveal a non-linear relationship between head mass and policy performance. The optimal head mass is 3.2 kg, achieving the best average performance (0.677), suggesting that moderate top-heaviness benefits the system by providing more pronounced proprioceptive signals while maintaining manageable inertial properties. Reducing mass to 2.2 kg presented challenges (0.660 average), as the controller struggled to develop a generalizable policy for  lighter heads.

Increasing the head mass beyond 3.2 kg consistently degraded performance: 4.2 kg dropped to 0.658, 5.2 kg fell significantly to 0.613, and 6.2 kg recovered slightly to 0.653 but remained well below optimal. This demonstrates that excessive mass creates inertial challenges that overwhelm the policy's compensation capabilities.
\subsubsection{Walking}

\begin{figure}[t]
    \centering
    \includegraphics[width=0.9\linewidth]{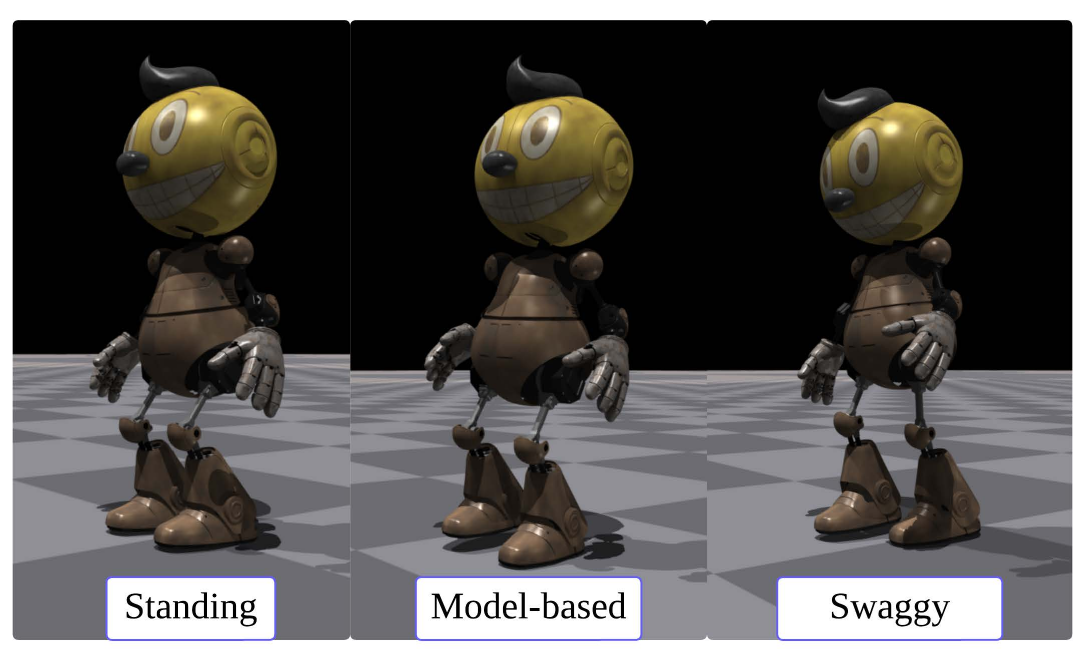}
    \caption{AMP policies with styles for balancing, model-based walking and walking with swagger.}
    \label{fig:training_results}
\end{figure}

As illustrated in Figure \ref{fig:training_results}, we developed three distinct policy styles: a basic standing pose, a model-based walking gait, and a more dynamic "swaggy" style incorporating natural human-like motion. 
These variations demonstrate our AMP framework's flexibility in capturing diverse movement aesthetics while maintaining stability.

Our walking policy evaluation focused on two critical parameters: reward structure weights and AMP style coefficient. Each row in Table \ref{tab:performance_metrics} represents a separately trained policy under the specified configuration, not evaluation of a single policy under different conditions. 
For a fair comparison of styles, a single baseline discriminator is used to evaluate the style rewards across experiments.
For reward structure, the balanced distribution [0.35, 0.35, 0.4] achieved the best average performance (0.721) with strong safety metrics (0.693). Alternative distributions like [0.5, 0.3, 0.2] improved individual task performance (0.757) but reduced safety (0.567).

For AMP style coefficient, a lower value (0.4) provided optimal overall performance (0.733) with excellent safety (0.720), while higher coefficients showed diminishing returns in safety: 0.7 achieved 0.729 average with 0.696 safety, and 0.9 reached 0.719 but degraded safety to 0.679. However, higher coefficients produced more human-like motion (discriminator loss improved from -0.420 to -0.372), highlighting the trade-off between robust performance and style fidelity for deployment.

\begin{table}[htbp]

\centering
\scriptsize % Reduce font size slightly
\caption{Performance metrics for balancing and walking tasks. Avg. represents the mean average of the previous 3 metrics. All results are averaged over 5 random seeds}
\setlength{\tabcolsep}{3pt} % Reduce column spacing further to accommodate ± values
\resizebox{\linewidth}{!}{
\begin{tabular}{lccccc}
\midrule
\multicolumn{6}{c}{\textbf{Balancing - Added Mass}} \\
\midrule
\textbf{kg} & \textbf{Motion} & \textbf{Task} & \textbf{Safety} & \textbf{Avg.} & \textbf{AMP} \\
\midrule
2.2 & 0.680±0.095 & 0.647±0.085 & 0.654±0.089 & 0.660±0.090 & -  \\
3.2 & 0.\textbf{693±0.091} & \textbf{0.663±0.085} & \textbf{0.675±0.090} & \textbf{0.677±0.089} & - \\
4.2 & 0.675±0.119 & 0.640±0.106 & 0.658±0.112 & 0.658±0.113 & - \\
5.2 & 0.631±0.092 & 0.594±0.081 & 0.615±0.087 & 0.613±0.087 & - \\
6.2 & 0.666±0.122 & 0.636±0.112 & 0.658±0.117 & 0.653±0.117 & - \\
\midrule
\multicolumn{6}{c}{\textbf{Walking - Reward Structure}} \\
\midrule
$[.35, .35, .4]$ & 0.723±0.107 & 0.743±0.118 & 0.693±0.109 & \textbf{0.721±0.111} & -0.392±0.036 \\
$[.5, .3, .2]$ & \textbf{0.773±0.113} & \textbf{0.757±0.123} & 0.567±0.091 & 0.699±0.110 & \textbf{-0.366±0.039} \\
$[.2, .6, .2]$ & 0.710±0.108 & 0.758±0.118 & 0.566±0.090 & 0.678±0.106 & -0.358±0.039 \\
$[.2, .4, .4]$ & 0.678±0.102 & 0.739±0.117 & \textbf{0.699±0.109} & 0.703±0.110 & -0.381±0.037 \\
\midrule
\multicolumn{6}{c}{\textbf{Walking - AMP Style Coefficient}} \\
\midrule
0.4 & \textbf{0.726±0.109} & \textbf{0.754±0.119} & \textbf{0.720±0.111} & \textbf{0.733±0.113} & -0.420±0.029 \\
0.5 & 0.722±0.109 & 0.749±0.118 & 0.715±0.112 & 0.720±0.113 & -0.412±0.029 \\
0.7 & 0.724±0.110 & 0.738±0.117 & 0.696±0.110 & \textbf{0.729±0.112} & \textbf{-0.372±0.033} \\
0.9 & 0.735±0.114 & 0.742±0.118 & 0.679±0.109 & 0.719±0.114 & -0.378±0.035 \\
\bottomrule
\end{tabular}
}
\label{tab:performance_metrics}
\end{table}

\subsection{ Sim-to-Real Transfer (Real Hardware)}
\begin{figure}[t]
    \centering
    \includegraphics[width=1.0\linewidth]{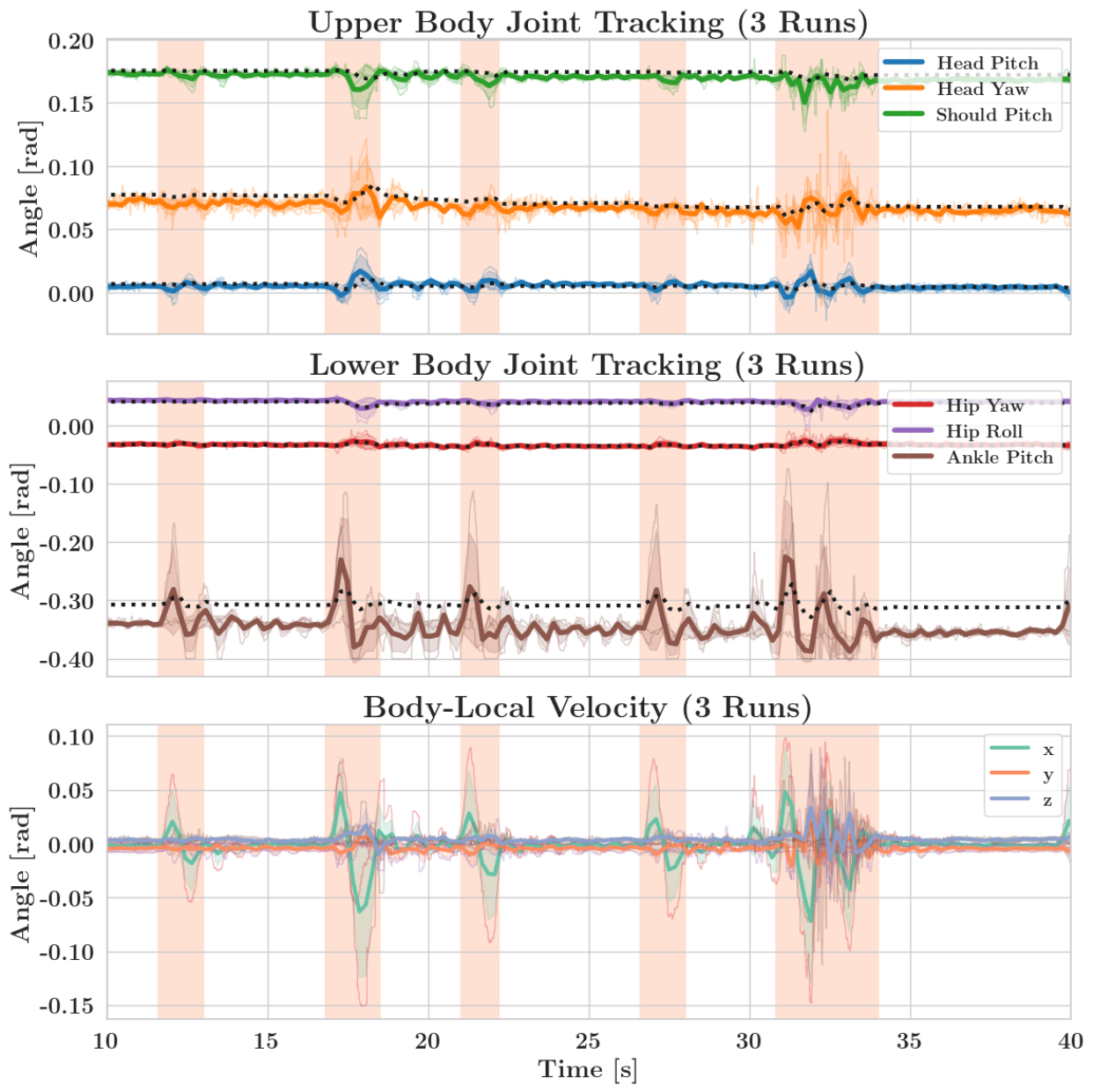}
    \caption{Joint tracking and disturbance rejection. Shaded regions indicate disturbance periods.}
    \label{fig:sim2real}
\end{figure}

Our AMP-trained policy achieves robust balancing capabilities as demonstrated in Figure \ref{fig:sim2real}. The upper body tracking data shows effective head stabilization despite its significant mass, with minimal deviations even during substantial perturbations up to 0.15 m/s.

Lower body tracking reveals sophisticated joint behaviors: Hip joints maintain stability while Ankle Pitch exhibits controlled oscillations during disturbance events (at 15s, 20s, and 30s), demonstrating that the policy has successfully learned human-like ankle-prioritized recovery strategies.

Body-local velocity data confirms strong disturbance rejection, with the system regaining stability within 2 seconds even after complex multi-directional perturbations. These results validate our domain randomization approach and reward structure.

\begin{figure}[t]
    \centering
    \includegraphics[width=1.0\linewidth]{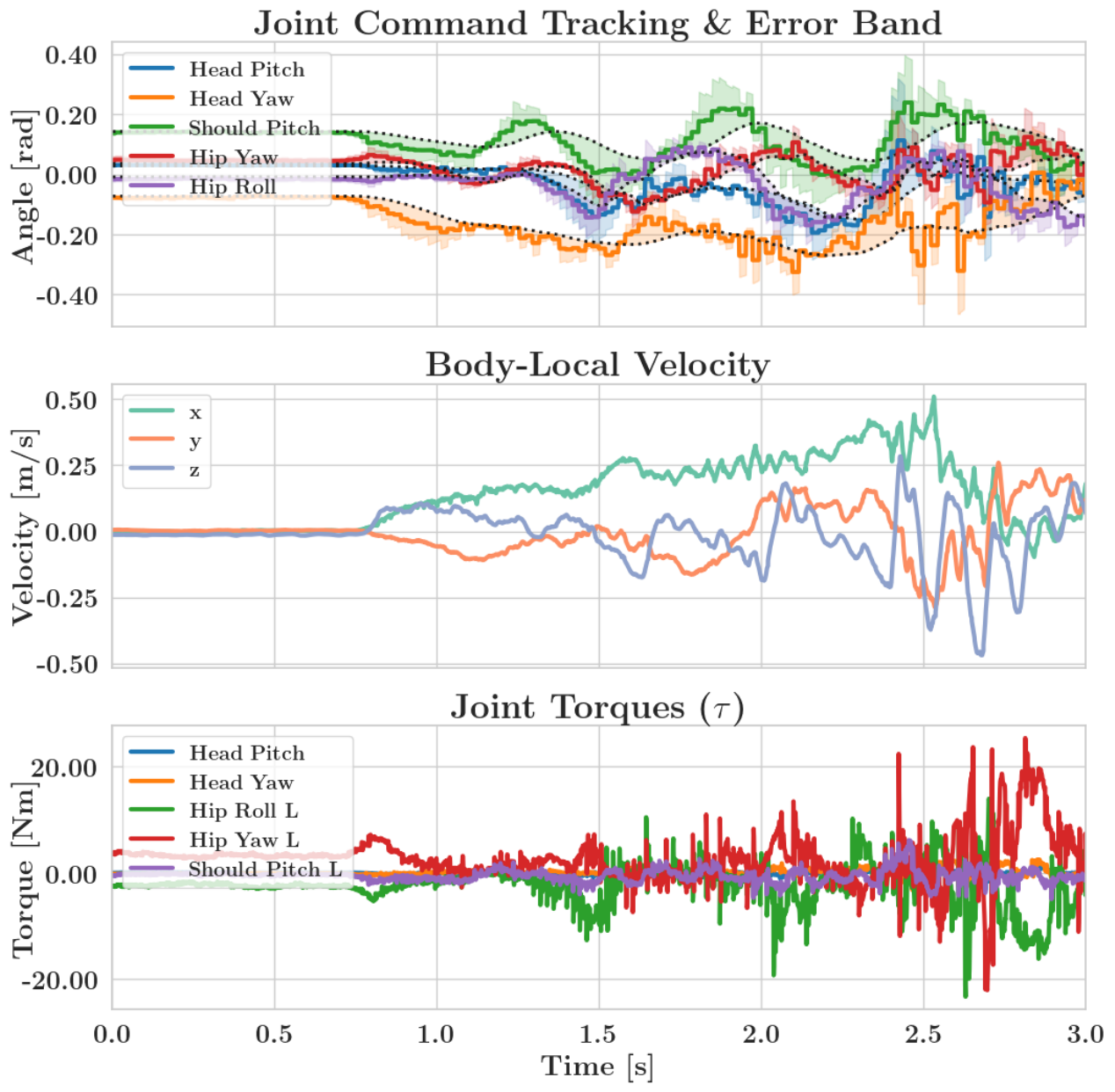}
    \caption{Joint command tracking, body-local velocity and joint torque tracking for natural walking.}
    \label{fig:walking_natural}
\end{figure}

The walking policy demonstrates successful transfer from simulation to hardware as shown in Figure~\ref{fig:walking_natural}. This natural walking implementation exhibits the dynamic, human-like characteristics that AMP is designed to produce. For instance, the policy reproduces key human movement patterns from the mocap data, including coordinated shoulder swing during stride and subtle head oscillations that create a more lifelike appearance. The data shows purposeful joint oscillations and pronounced velocity fluctuations in the x-direction, reaching peaks of 0.4 m/s commanded with distinctive patterns that authentically reproduce human walking gait cycles. This style prioritizes expressiveness, creating more engaging locomotion ideal for entertainment applications.

The policy successfully manages the robot's challenging mass distribution while maintaining joint torques within the safe operating range of $\pm 20$ Nm. Torque commands are clamped to actuator-specific torque limits, enforced in both simulation and hardware. The variable torque patterns during weight transfer phases reflect the sophisticated dynamics of this expressive gait.

These results validate our sim-to-real approach and demonstrate AMP's effectiveness in adapting to entertainment-focused designs while preserving natural human-like motion. Notably, the policy successfully respects hardware limitations and shell integrity constraints despite Cosmo's challenging morphology. Our successful hardware implementation confirms that our comprehensive domain randomization strategy effectively bridges the reality gap, even for robots with such demanding aesthetic and structural constraints.

\section{Ablation Studies}
\label{sec:ablation}
\subsection{Motion Styles}
\begin{figure}[ht!]
    \centering
    \begin{subfigure}[b]{0.31\textwidth}
        \centering
        \includegraphics[width=\textwidth]{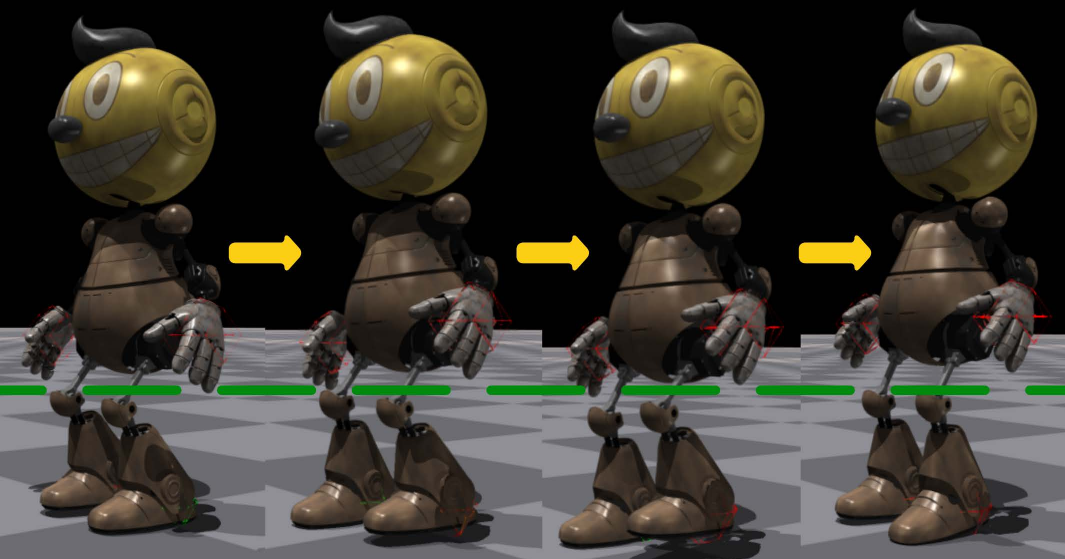}
        \caption{}
        \label{fig:cosmo_no_standing}
    \end{subfigure}
    \hfill
    \begin{subfigure}[b]{0.162\textwidth}
        \centering
        \includegraphics[width=\textwidth]{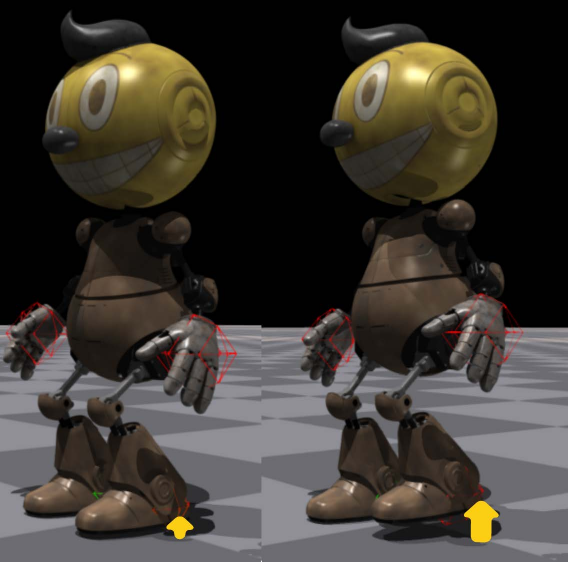}
        \caption{}
        \label{fig:cosmo_no_mb}
    \end{subfigure}
    \caption{Motion reference ablation: (a) Without standing reference; (b) Without model-based reference.}
    \label{fig:cosmo_ablation_studies_style}
\end{figure}

Our systematic ablation studies conclusively demonstrate the necessity of combining multiple reference types for achieving stable, natural locomotion, as shown in Figure \ref{fig:cosmo_ablation_studies_style}.

Without the standing pose reference, the robot exhibits dangerous vertical motion when transitioning to walking. Figure \ref{fig:cosmo_no_standing} reveals a nearly jumping behavior that significantly raises the center of mass above the reference line—dramatically increasing instability risk and energy consumption. This finding validates our methodical implementation approach from push recovery to a walking policy. 

More significantly, Figure \ref{fig:cosmo_no_mb} demonstrates that removing the model-based reference produces ineffective high-frequency stepping with poor foot trajectory control. Without this reference data guiding low-velocity transitions, the policy fails to accelerate from standstill to human-like walking speeds smoothly. The resulting gait exhibits rapid, inefficient foot placement without proper leg elevation. Importantly, our reward-only approach proves insufficient—proper foot trajectories require explicit reference motions rather than reward-based compensation alone.

\subsection{Reward Component Analysis}
\begin{figure}[t]
    \centering
    \includegraphics[width=1.0\linewidth]{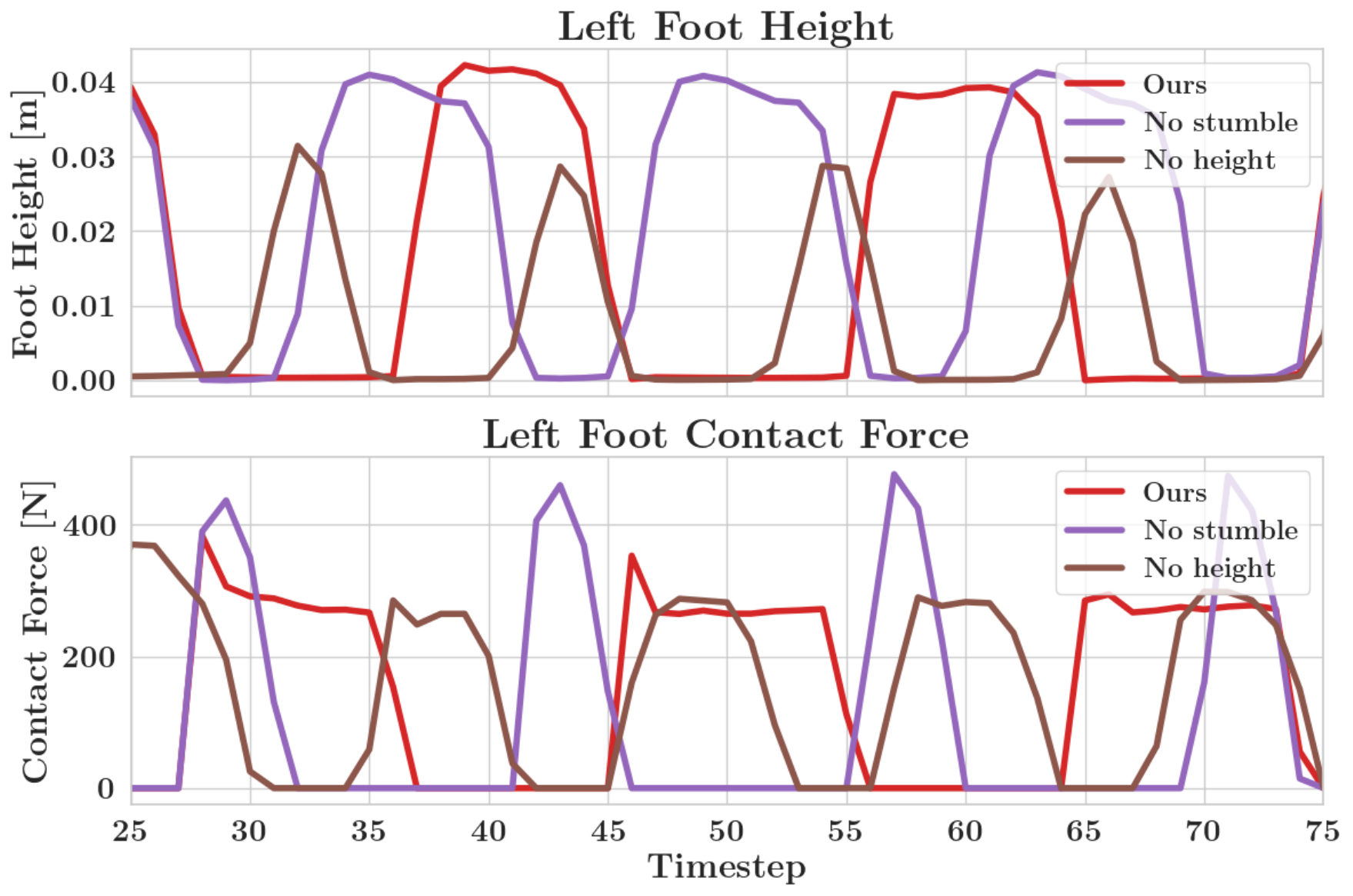}
    \caption{Foot height and contact force comparison under different reward configurations, illustrating the necessity of reward engineering for safety precautions}
    \label{fig:rewards_ablation}
\end{figure}

Our analysis establishes that specialized rewards beyond core imitation, motion quality, and task rewards are essential for managing foot trajectories and impact forces. Figure~\ref{fig:rewards_ablation} compares three configurations: our baseline approach (all rewards), variants without stumble prevention, and without foot height rewards.

The baseline configuration achieves controlled foot trajectories with moderate contact forces distributed over longer periods. At timestep 45, a complete baseline step takes approximately 12 control ticks, enabling gentle foot placement and preventing dangerous air phases. This translates to a biologically plausible gait frequency of approximately 1.9 Hz (based on 26 ticks × 0.02s per tick), which aligns well with natural human walking patterns. In contrast, configurations missing either height or stumble rewards produce faster but hazardous motions with dangerous impact force spikes, creating potential risks for hardware damage during deployment.

These specialized rewards protect the robot's physical components from damaging torque spikes during operation, enabling safer walking without compromising mobility.

\section{Discussion and Conclusion}
This work demonstrates the effectiveness of integrating physics-based simulation, adversarial motion priors, and domain randomization to enable locomotion in robots with challenging non-standard morphologies. Our successful implementation on Cosmo—with its disproportionately large head (16\% of total weight), complex collision geometries, and restrictive foot shells—confirms that AMP-guided reinforcement learning can adapt to aesthetic design constraints that defeat traditional approaches. 
This work demonstrates the viability of the AMP framework with eccentric morphologies, demonstrating behavioral adaptations—conservative ankle strategies and increased stabilization—while maintaining human-like motion style.

Key findings include: (1) specialized rewards are essential for hardware protection; (2) multiple motion references provide expressiveness and stability; and (3) domain randomization bridges reality gaps for unusual morphologies. While our approach prioritizes stability, platform safety, and natural motion over speed (achieving 0.5-0.7 m/s gaits), this trade-off aligns well with entertainment applications where expressiveness and hardware preservation often matter more than agility. 

Future work will compare imitation learning against traditional model-based controllers for entertainment robotics, where non-standard proportions and aesthetic constraints challenge conventional control approaches, and explore cross-embodiment transfer to validate broader applicability across diverse aesthetic morphologies.

\bibliographystyle{IEEEtran}
\bibliography{references}

\end{document}